\documentclass[letterpaper]{article} 
\usepackage{aaai25}  
\usepackage{times}  
\usepackage{helvet}  
\usepackage{courier}  
\usepackage[hyphens]{url}  
\usepackage{graphicx} 
\usepackage{natbib}  
\usepackage{caption} 
\usepackage{amsmath}
\usepackage{algorithm}
\usepackage{algorithmic}
\usepackage{booktabs}
\usepackage{subcaption}

\usepackage{newfloat}
\usepackage{listings}
\DeclareCaptionStyle{ruled}{labelfont=normalfont,labelsep=colon,strut=off} 
\floatstyle{ruled}
\newfloat{listing}{tb}{lst}{}
\floatname{listing}{Listing}
\title{ProcTag: Process Tagging for Assessing the Efficacy of Document Instruction Data}
\author{
    Yufan Shen\textsuperscript{\rm 1}\equalcontrib, 
    Chuwei Luo\textsuperscript{\rm 2}\equalcontrib\thanks{Corresponding Author}, 
    Zhaoqing Zhu\textsuperscript{\rm 2}\equalcontrib, 
    Yang Chen\textsuperscript{\rm 1},\\
    Qi Zheng\textsuperscript{\rm 2}, 
    Zhi Yu\textsuperscript{\rm 1}\footnotemark[2], 
    Jiajun Bu\textsuperscript{\rm 1},
    Cong Yao\textsuperscript{\rm 2}
}
\affiliations{
    \textsuperscript{\rm 1}Zhejiang Key Laboratory of Accessible Perception and Intelligent Systems, \\ College of Computer Science, Zhejiang University \\
    \textsuperscript{\rm 2}Alibaba Group

    {syficy,22351044,yuzhirenzhe,bjj}@zju.edu.cn \\
    {luochuwei,zzhaoqing.z,zhengqisjtu,yaocong2010}@gmail.com
}

\usepackage{marvosym}

\begin{document}

\maketitle

\begin{abstract}
  Recently, large language models (LLMs) and multimodal large language models (MLLMs) have demonstrated promising results on document visual question answering (VQA) task, particularly after training on document instruction datasets.
  An effective evaluation method for document instruction data is crucial in constructing instruction data with high efficacy, which, in turn, facilitates the training of LLMs and MLLMs for document VQA.
  However, most existing evaluation methods for instruction data are limited to the textual content of the instructions themselves, thereby hindering the effective assessment of document instruction datasets and constraining their construction.
  In this paper, we propose ProcTag, a data-oriented method that assesses the efficacy of document instruction data.
  ProcTag innovatively performs tagging on the execution process of instructions rather than the instruction text itself.
  By leveraging the diversity and complexity of these tags to assess the efficacy of the given dataset, ProcTag enables selective sampling or filtering of document instructions.
  Furthermore, DocLayPrompt, a novel semi-structured layout-aware document prompting strategy, is proposed for effectively representing documents.
  Experiments demonstrate that sampling existing open-sourced and generated document VQA/instruction datasets with ProcTag significantly outperforms current methods for evaluating instruction data.
  Impressively, with ProcTag-based sampling in the generated document datasets, only 30.5\% of the document instructions are required to achieve 100\% efficacy compared to the complete dataset.
\end{abstract}

%

\section{Introduction}
\label{sec:intro}


Document visual question answering (VQA), which closely aligns with the general document artificial intelligence~\cite{cui2021document,han2023workshop}, is currently a significant research and application area.
With the remarkable successes achieved by Large Language Models (LLMs)~\cite{chatgpt_webpage,gpt3,touvron2023llama,touvron2023llama2,bai2023qwen} and Multimodal Large Models (MLLMs)~\cite{openai2023gpt4,bai2023qwenvl,liu2023llava,liu2023improvedllava,ye2023mplugowl} across various real-world tasks, the application of these models~\cite{zhang2023llavar,ye2023mplugdoc,yang2023dawn,shi2023exploring,liu2023hidden} for the document VQA has seen substantial advancements.

To achieve good document VQA results by LLMs/MLLMs, recent works~\cite{ye2023mplugdoc,zhang2023llavar} perform instruction tuning~\cite{wei2021finetuned,instructgpt} on the document instruction datasets.
To facilitate effective instruction tuning, the instruction datasets should better possess high efficacy, which means that the instructions in the dataset are expected to be sufficiently diverse and complex~\cite{xu2023wizardlm,mukherjee2023orca,wang2024far}.
Instruction datasets are often collected through manual annotation or generated using LLMs/MLLMs~\cite{alpaca,ding2023enhancing}.
Building upon this, recent works~\cite{lu2023instag,li2023quantity} have proposed several data-oriented methods to evaluate the efficacy of instruction datasets, thereby facilitating the selection of data that is both diverse and complex.
These methods have been proven effective in guiding the assessment of instruction data quality.


\begin{figure}[tp]
  \centering
  \includegraphics[width=.99\linewidth]{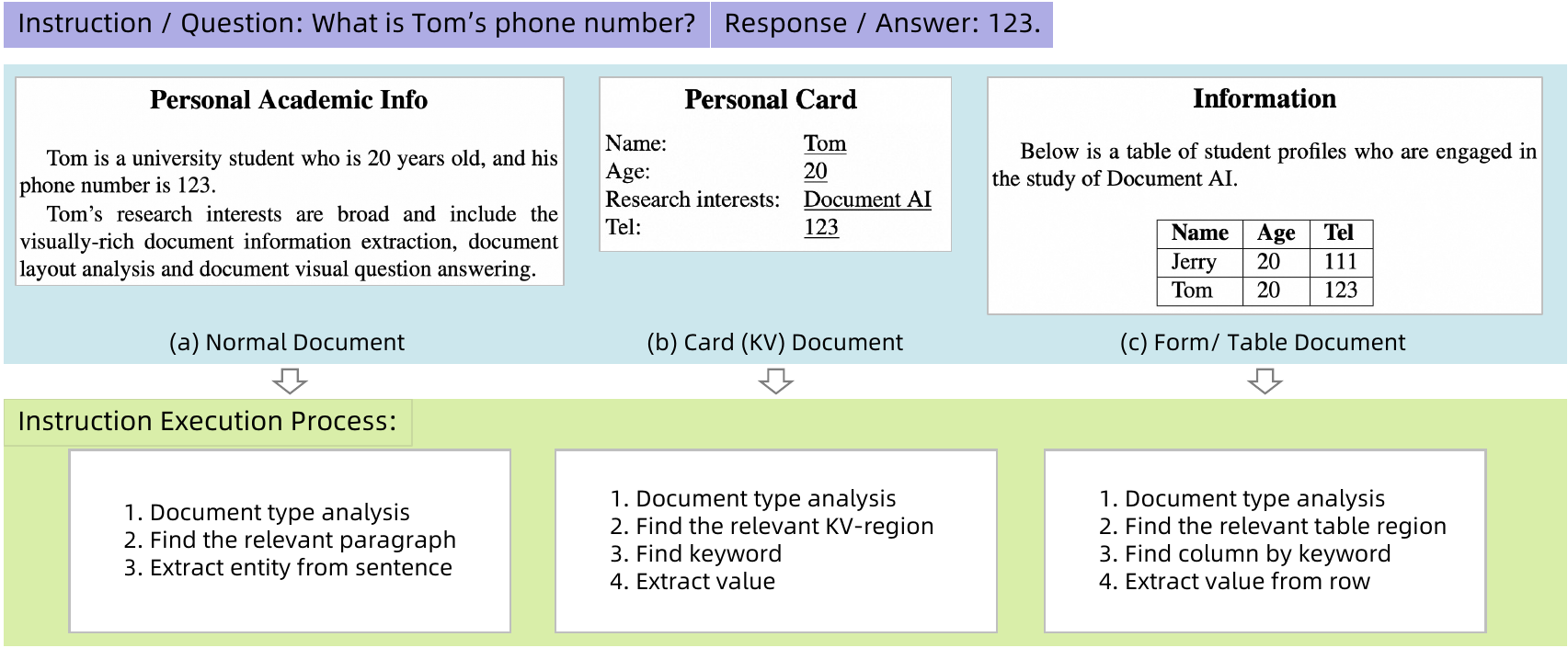}
  \caption{The same instruction text can lead to entirely different execution processes when applied to different types of documents.}
  \label{fig:intro}
\end{figure}

However, existing evaluation methods for instruction data are biased to the general domain and primarily focus on the textual content of the instructions, resulting in suboptimal performance in the document understanding domain.
In the document VQA task, effectively assessing the efficacy of instruction datasets requires a comprehensive understanding and differentiation of the document's content, layout, instructions, and the \textit{instruction execution process}.
As illustrated in Figure~\ref{fig:main_result}, the data method INSTAG~\cite{lu2023instag}, which assesses the efficacy of instruction datasets by the diversity and complexity of instruction text, is applied to DocVQA datasets for training LLMs.
The performance of LLMs trained on this filtered dataset exhibits minimal difference compared to models trained on randomly sampled data.
The primary issue lies in the fact that, in document understanding, the same instruction text may lead to entirely different data categories depending on the instruction execution processes.
Data methods that only focus on the instruction text are incapable of effectively distinguishing these differences.
Consider the instruction ``What is Tom's phone number?" as shown in Figure~\ref{fig:intro}. 
When evaluated using methods like INSTAG, which considers only the text of the instruction, this instruction would be tagged as ``information extraction/entity extraction".
In the document understanding scenarios of Figure~\ref{fig:intro} (a), (b), and (c), it is evident that the same instruction carries three entirely different instruction execution processes.
In (a), it is required to locate and understand a paragraph within the document; in (b), the focus shifts to identifying key-value regions before comprehension; and in (c), it involves finding the relevant table region for table understanding. Clearly, the same instruction text can represent entirely different instruction types in different document scenarios. However, there’s a significant lack of research on evaluating the effectiveness of document instruction data, which hinders the development of high-quality datasets for training LLMs/MLLMs in document VQA. Therefore, exploring effective methods for assessing document instruction datasets is crucial.

To this end, we propose ProcTag, a metric for assessing the efficacy of document instruction data by instruction execution process tagging.
ProcTag focuses on modeling the process of document instruction execution rather than the instruction text itself. 
Empowered by the high-performing LLM GPT-3.5, including its chain-of-thought~\cite{wei2022chain,kojima2022large} reasoning capability, ProcTag introduces a framework that prompts GPT-3.5 to generate the process of document instruction execution.
To better represent the process, ProcTag requires GPT-3.5 to express the process in code and then assign tags to the generated process. These tags are used to measure the diversity and complexity, representing the efficacy of the instruction data.
Furthermore, to comprehensively represent document content and utilize GPT-3.5 more effectively, we also devise DocLayPrompt, a novel semi-structured prompting strategy based on document layout information for document representation.
Compared to existing layout-aware prompts~\cite{lamott2024lapdoc}, DocLayPrompt outperforms them in capturing the layout information for document representation.
Given the relative scarcity of document instruction datasets, we apply the ProcTag method to an existing open-source document VQA dataset and several generated document instruction datasets. 
After applying ProcTag for data selection, the datasets are used to train both LLMs and MLLMs for experiments. 
Experimental results demonstrate the effectiveness of the proposed ProcTag method in assessing document instruction data with efficacy when compared to existing data methods and random sampling.

Our contributions are summarized as follows:
\begin{itemize} 
  \item[1)] This paper introduces a framework ProcTag,  which models the instruction execution process instead of the instruction text itself, as a method for assessing the efficacy of document instruction data. To the best of our knowledge, it is the first to explore the instruction execution process as a data quality assessment method for document understanding.
  \item[2)] To effectively model the instruction execution process, ProcTag expresses it for document VQA problems in code and distinguishes the process through tagging. Additionally, a semi-structured layout prompting strategy named DocLayPrompt, which incorporates document layout information for effective document representation, is proposed.
  \item[3)] The proposed ProcTag method, when applied to existing open-source document VQA dataset and generated document understanding instruction datasets, significantly outperforms existing data methods when training LLMs and MLLMs. Additionally, by guiding the generation and filtering of document understanding instruction datasets using ProcTag, only 30.5\% of the document instructions are required to achieve 100\% efficacy compared to the complete dataset.
\end{itemize}

\section{Related Works}
\noindent{\bf{LLMs/MLLMs for Document VQA.}}
Recently, there have been broad discussions within the community regarding LLMs such as ChatGPT~\cite{chatgpt_webpage}, LLaMA~\cite{touvron2023llama,touvron2023llama2} and Qwen~\cite{bai2023qwen}, and MLLMs like GPT-4V~\cite{openai2023gpt4}, Gemini~\cite{team2023gemini} and Qwen-VL~\cite{bai2023qwenvl}. These models have achieved considerable success in a wide broad of downstream AI applications. Concurrently, leveraging LLMs/MLLMs for document AI~\cite{yang2023dawn,liu2023hidden,perot2023lmdx,shi2023exploring}, especially document VQA tasks~\cite{zhang2023llavar,ye2023mplugdoc,bai2023qwenvl}, has demonstrated remarkable performance over previously document pre-trained models~\cite{xu2020layoutlm,xu2020layoutlmv2,huang2022layoutlmv3,li2021selfdoc,luo2022bivldoc,li2021structext,appalaraju2021docformer,gu2022unified,peng2022ernielayout,Luo_2023_CVPR,xing2023lore}.
LLaVAR~\cite{zhang2023llavar} enhances the text comprehension capabilities of LLaVA~\cite{liu2023llava} by gathering text-rich images and constructing a corresponding instruction tuning dataset.
Moreover, building upon the foundation of mPLUG-Owl~\cite{ye2023mplugowl}, mPLUG-DocOwl~\cite{ye2023mplugdoc} creates an instruction tuning dataset for various visual-text understanding tasks. Simultaneously, an OCR-free document instruction comprehension evaluation set, LLMDoc, has been developed to better compare models in terms of instruction compliance and document understanding. 
Likewise, Qwen-VL~\cite{bai2023qwenvl} also considers utilizing high-quality and fine-grained visual-language (VL) instruction datasets to achieve its high-quality multimodal multitask understanding capabilities, which also includes instruction fine-tuning datasets related to documents.
LayoutLLM~\cite{luo2024layoutllm} proposes a layout instruction tuning method and a document instruction tuning dataset for document understanding. A module called layout chain-of-thought (LayoutCoT) which is effective for document understanding is devised to represent the document-related instruction processes.
The success of the aforementioned efforts demonstrates that, for document-related tasks, the document instruction datasets are indispensable.

\noindent{\bf{Instruction Data Methods.}}
%
Instruction tuning datasets~\cite{wei2021finetuned,instructgpt} are crucial for calibrating LLMs/MLLMs to align human instructions accurately. The majority of these datasets are sourced by manual annotation or generated by LLMs and MLLMs~\cite{alpaca,ding2023enhancing}.
Recent efforts in the community have recognized the importance of establishing robust instruction dataset evaluation methods to optimize the utility of these datasets \cite{li2023quantity,lu2023instag}.
Li et al. \cite{li2023quantity} introduce the Instruction-Following Difficulty (IFD) metric, which autonomously screens for high-quality instruction data by identifying discrepancies between a model's expected responses and its generative outputs. Despite its innovative approach, this method is hindered by the need for additional training on a pre-experienced model.
In contrast, InsTag \cite{lu2023instag} proposes a more cost-efficient alternative that foregoes the need for model retraining. By leveraging existing LLMs to annotate instruction data, InsTag assesses quality along two dimensions: diversity and complexity. This approach also highlights that the quality of instruction datasets is crucial for impacting model performance. 
For document AI tasks, publicly available document instruction datasets are scarce. Furthermore, the above evaluation metrics are biased towards the general domain and focus primarily on textual instructions. These metrics can not be directly applied to complex multimodal tasks like document VQA, which relies not only on instruction text but also on the understanding of layout and different problem-solving processes. 
To more effectively employ document instruction datasets and enhance the document understanding capability of LLMs/MLLMs, it is necessary to develop an evaluation metric suitable for assessing instruction data in document VQA.

\section{ProcTag}

\begin{figure*}[tp]
  \centering
  \includegraphics[width=.99\linewidth]{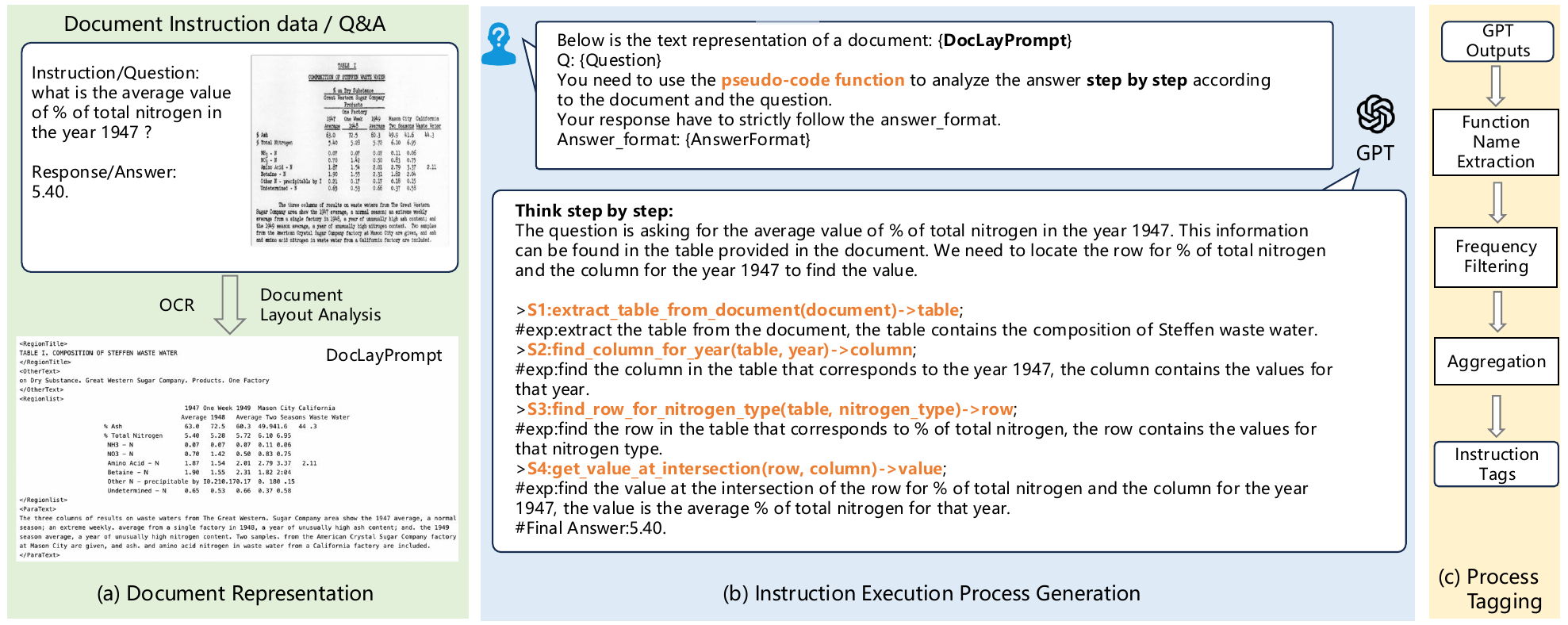}
  \caption{Overview of ProcTag. ProcTag performs tagging on the document instruction execution process for assessing the efficacy of document instruction data, involving three steps: (a) document representation: ProcTag utilizes DocLayPrompt for representing document information; (b) instruction execution process generation: prompting GPT to generate the execution process using pseudo-code; and (c) process tagging: processing the generated pseudo-code to obtain instruction tags.}
  \label{fig:overview}
\end{figure*}
Leveraging tags to measure the diversity and complexity for assessing the efficacy of instruction datasets is a very effective way~\cite{lu2023instag}.
As shown in Figure~\ref{fig:overview}, ProcTag tags document instruction datasets by modeling the instruction execution process, including 3 steps.

\subsection{Document Representation}
\label{sec:Document Representation}

Effectively representing document content with rich layout information is crucial for tagging document instruction datasets. While directly inputting document images into Multimodal Large Language Models (MLLMs) like GPT-4V is a straightforward approach, it is both costly and inefficient. Research has shown that textual representation alone can achieve satisfactory results in document understanding tasks~\cite{lamott2024lapdoc}. Methods such as LATIN-Prompt~\cite{wang2023layout} and SpatialFormat-prompt~\cite{lamott2024lapdoc} reconstruct layout information using Optical Character Recognition (OCR) results, spaces, and line breaks. However, these approaches focus solely on text content, neglecting layout details, which can lead to ambiguities. For instance, it becomes challenging to distinguish between elements like \textit{Page Head} and \textit{Title} when only text is considered, as they often appear at the top of the page with similar fonts.

To address this limitation, we propose a novel semi-structured approach called DocLayPrompt, which incorporates layout information into document representation, enhancing the tagging of document instruction datasets. As shown in Figure~\ref{fig:overview} (a), DocLayPrompt supports comprehensive modeling of a document’s content by integrating layout details.

In DocLayPrompt, an input document image $\mathcal{D}$ undergoes OCR and layout detection to extract structural information: OCR results $\mathcal{O} =  <\mathcal{S}, \mathcal{B}O>$ and layout results $\mathcal{L} = <\mathcal{T}, \mathcal{B}_L>$. Specifically, $\mathcal{S} = \{s_1, s_2, \ldots, s_n\}$ represents the extracted text sequences, and $\mathcal{B}_O = \{b_{o1}, b_{o2}, \ldots, b_{on}\}$ corresponds to their bounding box coordinates. Simultaneously, $\mathcal{T} = \{t_1, t_2, \ldots, t_n\}$ denotes the identified layout types, with $\mathcal{B}_L = \{b_{l1}, b_{l2}, \ldots, b_{ln}\}$ specifying the bounding box coordinates for these layout components. These elements are then utilized to construct DocLayPrompt.

\noindent 1) \textbf{Reorder and Clean Inputs}: 
To represent real documents accurately, it’s crucial to use clean and organized structural information, such as OCR and layout results. This involves reordering these outcomes using their bounding box coordinates, $\mathcal{B}_O$ and $\mathcal{B}_L$. Redundant layout instances are removed using Non-Maximum Suppression (NMS), leading to cleaned inputs: $\mathcal{O}^C$ and $\mathcal{L}^C$.
\begin{align}
  \mathcal{O}^C, \mathcal{L}^C &\leftarrow \textit{getCleanInputs}(\mathcal{O}, \mathcal{L}).
\end{align}

%
\noindent 2) \textbf{Associate Structural Information}:
Both $\mathcal{O}^C$ and $\mathcal{L}^C$ represent the structural information within documents. To create a unified representation, these sources need to be combined into associated structural information, $\mathcal{A}{OL}$. Since $\mathcal{L}^C$ covers a broader structural granularity than $\mathcal{O}^C$, DocLayPrompt integrates a set of OCR results ${\mathcal{O}^C_i}^n{i=m}$ into their corresponding layout component $\mathcal{L}^C_j$ for a tight association. If an OCR result $\mathcal{O}^C_i$ is not encompassed by any layout component, it is associated with the nearest layout component $\mathcal{L}^C_j$ based on Euclidean distance $d(\cdot)$.


\begin{equation}
\begin{split}
  & \hspace{2cm} \mathcal{A}_{OL} \leftarrow \{(\{\mathcal{O}^C_i\}^n_{i=m}, \mathcal{L}^C_j)\}, \\
  & \begin{array}{c}
    \text{where } \{\mathcal{B}_{Oi}\}^n_{i=m} \subseteq \mathcal{B}_{Lj} \text{ or } \underset{i, j}{\arg\min} \, d(\{\mathcal{B}_{Oi}\}^n_{i=m}, \mathcal{B}_{Lj}).
  \end{array}
\end{split}
\end{equation}

%

\noindent 3) \textbf{Represent Document}:
Finally, concatenate $\mathcal{A}_{OL}$ sequentially through LATIN-Prompt and layout type tags to form the document representation $\mathcal{R}$.
\begin{equation}
  \mathcal{R} \leftarrow \textit{getDocRep}(\mathcal{A}_{OL}).
\end{equation}

\subsection{Instruction Execution Process Generation}
\label{sec:Instruction Execution Process Generation}

As shown in Figure~\ref{fig:intro}, effectively distinguishing document instruction data requires focusing on the instruction execution process rather than just the textual content. Inspired by the chain-of-thought (CoT)~\cite{wei2022chain,kojima2022large} capabilities of LLMs, ProcTag uses GPT to generate the document instruction execution process. To make this process more precise and concise, ProcTag represents it in pseudo-code. For greater accuracy, GPT first outputs a CoT textual description of the instruction execution, which is then used to generate the corresponding pseudo-code.

As shown in Figure~\ref{fig:overview} (b), the structural representation $\mathcal{R}$ and corresponding question $\mathcal{Q}$ for the document instruction dataset are employed to prompt GPT to generate instruction execution process content $\mathcal{P}$.
To ensure sufficient and clear discriminability for subsequent tagging, the generated $\mathcal{P}$ includes a chain-of-thought (CoT) for instruction execution and the corresponding tightly coupled and distinct pseudo-code execution process, where the input of each step is the output of the previous step.

\subsection{Process Tagging}
\label{sec:Process Tagging}


To obtain distinctive and denoised process tags for assessment, ProcTag normalizes process tags through 3 stages: \textit{Function Name Extraction}, \textit{Frequency Filtering}, and \textit{Aggregation}. As shown in Figure~\ref{fig:overview} (c), \textit{Function Name Extraction} is utilized to obtain the function names of pseudo-code as the raw process tags from $\mathcal{P}$.
Then, following the InsTag, the raw process tags employ \textit{Frequency Filtering} and \textit{Aggregation} to filter tags that appear too rarely and to aggregate similar tags (e.g., ``find\_table'' and ``extract\_table''), respectively.
To filter out tags that appear infrequently, a threshold is established for their selection. To effectively aggregate semantically similar tags with code-like formats, a code language model is utilized to obtain embeddings for these code-formatted tags, which are then clustered.
After that, the final instruction tags obtained through the above stages will be used for subsequent data assessment and experiments.

\subsection{Efficacy Assessment by Tags}
\label{sec:Efficacy Assessment by Tags}

ProcTag defines the efficacy of an instruction dataset through two attributes of its process tags: complexity and diversity. Here, complexity refers to the number of different tags present within the dataset, while diversity denotes the average number of tags per instruction data. Higher levels of complexity and diversity signify a dataset with higher efficacy, whereas lower levels indicate lower efficacy.
Following the InsTag data method for data sampling via tags, the ProcTag selects subsets by optimizing for the highest complexity under the condition of maximal tag diversity, thereby yielding sub-datasets with superior efficacy.


\begin{table}[tb]
  \centering
  \small

  \begin{tabular}{lcccc}
  \toprule
  \textbf{Metric Agreement} & \textbf{GPT-4} & \textbf{Human} & \textbf{Human-GPT} \\
  \midrule
  \textbf{Tag Precision} & 96\% & 92\% & 0.65 \\
  \textbf{Tag Consistency} & 80\% & 88\% & 0.87  \\
  \bottomrule
  \end{tabular}
  \label{tab:quality_evaluation}
  \caption{
    The quality evaluation of the tags generated by ProcTag. Tagging precision and consistency are utilized for evaluating ProcTag. The Cohen's kappa score is used to represent the agreement between GPT-4 and human.
  }
\end{table}

\section{Experimental Setup}
\subsection{Datasets}
\label{sec:Datasets}
Instruction datasets, including document instruction datasets, are often collected from manual annotations or generated by LLMs or MLLMs.
In our experiments, all these types of document instruction datasets are considered for tagging and evaluation.

\noindent{\textbf{Manually Annotated Dataset.}}
The widely-used public dataset \textbf{DocVQA}~\cite{mathew2021docvqa} is employed, consisting of 50,000 questions defined across over 12,000 documents from various industry sources.


\noindent{\textbf{Generated Dataset.}}
For document VQA tasks, the availability of publicly accessible instruction datasets is limited. Therefore, motivated by existing works that utilize LLMs/MLLMs for generating instruction datasets~\cite{alpaca,ding2023enhancing}, it is necessary to use LLMs/MLLMs to generate some expanded corpus of document instruction datasets.
Motivated by the LayoutLLM\cite{luo2024layoutllm}, to cover a broader spectrum of document types and complexity instructions, four document datasets are used for building the instruction tuning dataset, including \textbf{RVL-CDIP}~\cite{harley2015evaluation}, \textbf{DocILE}~\cite{2023docile}, \textbf{PublayNet}~\cite{zhong2019publaynet}, and \textbf{PubTabNet}~\cite{zhong2020image}.
RVL-CDIP covers a substantial diversity of document types, with a collection of 400,000 images spanning  16 distinct classes, including, but not limited to, letters, forms, and memos.
DocILE is specialized in form-based documents that facilitate diverse information extraction tasks. It contains a set of 6,680 labeled business documents, supplemented by a substantial unlabeled set of 932,000 documents, and a synthetically generated corpus of 100,000 documents.
PublayNet, with its focus on document layout analysis, comprises over 360,000 PDF documents rich in textual and layout elements. This dataset features annotations that include common layout components such as text blocks, titles, lists, figures, and tables.
Lastly, PubTabNet is characterized by its intricate table structures contained within academic literature. Encompassing more than 568,000 tabular instances in both image and HTML formats, this dataset offers detailed cell bounding box information, which is instrumental for advanced table recognition and understanding tasks.

\begin{figure}[tb]
  \centering
  \includegraphics[width=0.6\linewidth]{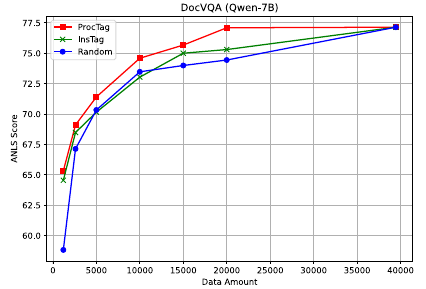}
  \includegraphics[width=0.8\linewidth]{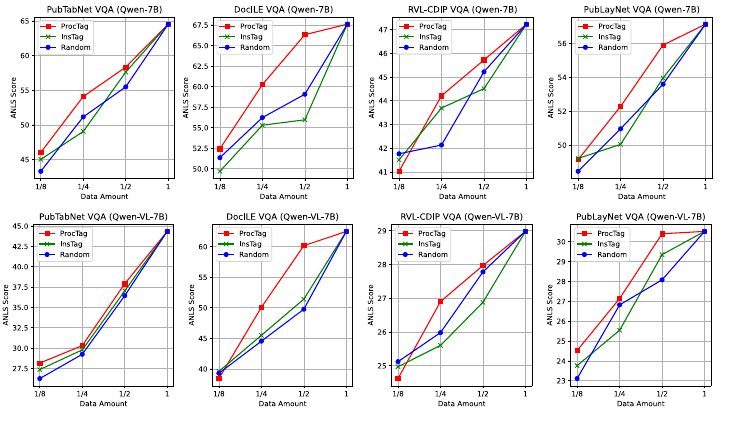}
  \caption{Experimental results of the performance in document VQA after training on LLM (Qwen) and MLLM (Qwen-VL) using datasets sampled with ProcTag, InsTag, and random sampling methods from human-annotated (DocVQA) and generated document instruction datasets.
  }
  \label{fig:main_result}
\end{figure}

\begin{figure}[tb]
  \centering
  \includegraphics[width=0.8\linewidth]{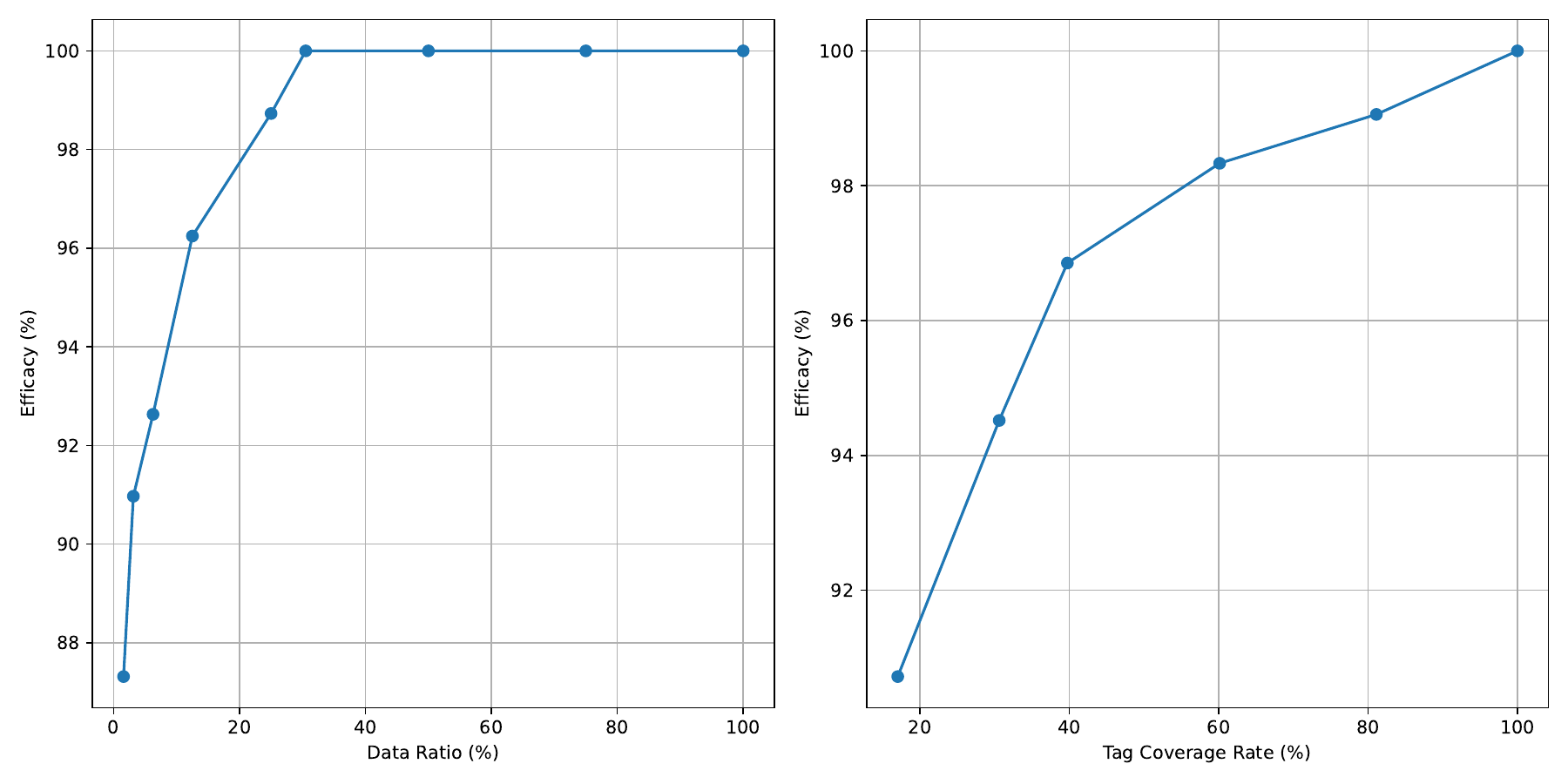}
  \caption{
  Experimental analysis results of data efficacy in terms of data amount ratio and tag coverage rate.
  }
  \label{fig:efficiency_200k}
\end{figure}

\subsection{Implementation Details}

\textbf{ProcTag Implementation Details.}
(1) \textit{Document Representation.} 
The Vision Grid Transformer (VGT)~\cite{da2023vision} model is utilized as the layout detection tool described in the section Document Representation for structural document representation.
The OCR results for document images are sourced from the official OCR outputs provided by each respective document dataset.

(2) \textit{Instruction Execution Process Generation.}
During the instruction execution process generation in the section Instruction Execution Process Generation, if the output format returned from GPT does not meet the generation criteria and cannot be parsed, GPT is invoked again. This process is repeated at most twice, and if it still fails, the data is discarded. The proportion of discarded samples does not exceed 0.1\%.

(3) \textit{Process Tagging.}
In the section Process Tagging, as following InsTag, the tags appearing fewer than 4 times in DocVQA and fewer than 2 times in other datasets are termed as long-tail tags, removing them in ``Frequency Filtering".
During the ``Aggregation" stage, after the exclusion of lowercase letters and special characters, embeddings for function names are obtained using the SantaCoder~\cite{allal2023santacoder} and clustered with DBSCAN~\cite{hahsler2019dbscan}. The minimum semantic similarity threshold is set to 0.015. 
Finally, a minimum support of 40 instances and a minimum confidence of 99\% are used for association aggregation. 
It is important to note that for instruction execution processes, the same types of tags with different orders represent different execution processes.
Hence, ProcTag aggregates tags only when they exhibit a high frequency of occurrence and are in close proximity to the execution order.
For example, ``extract\_list" and ``find\_item" are commonly co-occurring, but if they respectively appear in the first and third steps of the execution process that are non-adjacency, they are not considered to be aggregated. Conversely, they would be combined to ``extract\_list\_item".

\noindent{\textbf{Dataset Details.}}
\textit{Instruction Generation.}
PubLayNet, PubTabNet, RVL-CDIP, and DocILE are utilized for instruction generation.
For each dataset, 3,000 document images are randomly sampled to prompt GPT (GPT-3.5) with the same prompt that is used for generating the layout instruction tuning dataset of the LayoutLLM\cite{luo2024layoutllm}.
10 instruction items are generated for each document. 
After parsing and filtering, a random selection of 20,000 instances forms the training set, with another 1,000 instances randomly selected to serve as the test set.

\noindent{\textbf{Experimental Details.}}
\textit{LLMs and MLLMs.}
Given the superior performance of the Qwen series models on the DocVQA task within open-source models, we choose the Qwen series models as our experimental benchmark.
Both LLMs and MLLMs are employed for experimental validation.
For LLMs, all experiments are based on the 7B version of QwenChat. 
For each experiment, the models are trained for 3 epochs with a batch size of 32.
As for MLLM models, the 7B version of QwenVL-Chat is employed, trained for 3 epochs, with a batch size of 16.
For all experiments, the learning rate is set to 1e-5.

\noindent{\textbf{Evaluation Metric.}}
The widely used metric ANLS~\cite{mathew2021docvqa} is utilized for evaluating DocVQA and the four generated document instruction datasets.

\subsection{Quality Evaluation}

Following the InsTag method, the quality of the tags generated by ProcTag is evaluated from the perspectives of precision and consistency.
Precision is defined to measure whether tags correctly relate to the execution processes of their associated instructions, while consistency assesses whether the tags maintain a uniform definition across all corresponding instructions.
GPT-4 and manual annotation are utilized to evaluate the tags generated by ProcTag.

\textbf{Results.} Table~\ref{tab:quality_evaluation} shows the quality evaluation of the tags generated by ProcTag in the DocVQA dataset.
The precision of our ProcTag on GPT-4 and human annotation reaches 96\% and 92\%, respectively, with consistency both exceeding 80\%. 
Furthermore, Cohen's kappa score is used to compute the agreement between human and GPT-4 annotation, yielding precision and consistency agreements exceeding 0.6, which qualifies as ``substantial agreement". Moreover, an additional comparison of final answers derived from step-by-step reasoning via ProcTag depicted in Figure~\ref{fig:overview}(b) with ground truth shows an ANLS increase from 73.84 to 78.68 compared to direct reasoning without ProcTag. 
In conclusion, the tags marked by ProcTag demonstrate high quality in both precision and consistency.

\section{Experimental Results}

\subsection{Main Results}

Our experiments are conducted on the widely-used open-source DocVQA dataset which is manually annotated, as well as on four generated datasets, for supervised fine-tuning of LLMs/MLLMs.
For each dataset, models are fine-tuned with different data proportions sampled by three methods: ProcTag, the existing instruction data method InsTag, and random sampling. 
Then, the document understanding performance of models in these states is evaluated on the test set, where Qwen and Qwen-VL are used as models.
As shown in Figure~\ref{fig:main_result}, overall, our ProcTag consistently outperforms both InsTag and random variants. Notably, the performance trends of InsTag and random methods are similar. This is because InsTag, which only considers instruction text, cannot effectively distinguish different document instruction data, this finding is consistent with our observations (see Figure~\ref {fig:intro}), which proves the necessity of modeling the process of document instruction execution rather than the instruction text itself.

\noindent{\textbf{Evaluation on Manual Annotation Dataset.}} Due to the scarcity of existing document instruction datasets, Qwen-VL has been trained on most datasets of this category, including DocVQA. 
For fair comparison, our experiments avoid validating on Qwen-VL, considering only Qwen.

\noindent{\textbf{Evaluation on Generated Dataset.} }
The experimental results corresponding to the four generated datasets show consistent trends on both Qwen and Qwen-VL. 
In particular, unstable performance has been observed in the setting with 1/8 data proportion on certain datasets (like DocILE and RVL-CDIP), which we hypothesize is caused by limited data amount.
Moreover, it can be seen that the performance achieves the highest on DocILE, attributable to its highly diverse document content. Conversely, ProcTag performs the least effectively on PubTabNet and PubLayNet, which are confined to tabular and academic paper layouts, respectively. The performance on RVL-CDIP is among them.
In summary, our ProcTag exhibits superior performance on datasets with more diverse document data.

\begin{figure}[htbp]
  \centering
  \begin{minipage}[b]{0.23\textwidth}
    \includegraphics[width=\textwidth]{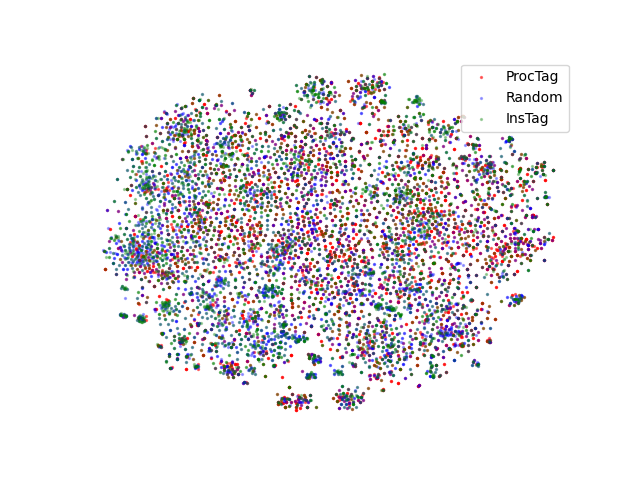}
    \subcaption{image features}
    \label{fig:leftimage}
  \end{minipage}
  \hfill 
  \begin{minipage}[b]{0.23\textwidth}
    \includegraphics[width=\textwidth]{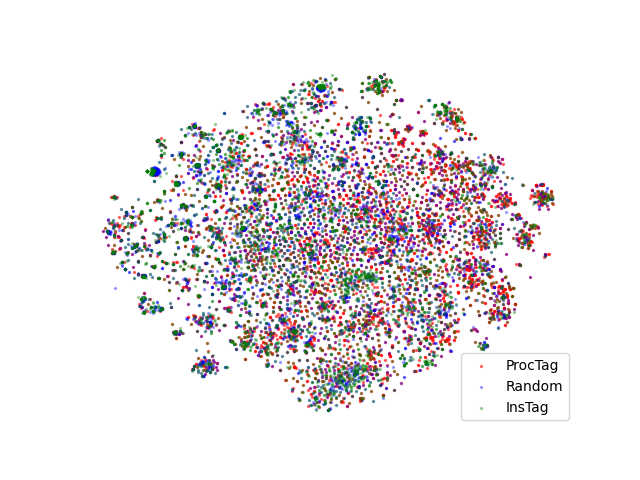}
    \subcaption{instruction text features}
    \label{fig:rightimage}
  \end{minipage}
  \caption{
  t-SNE visualization of the document image and instruction text features obtained by sampling DocVQA trainset with three methods (ProcTag, Random, and InsTag) using the CLIP feature extractor.
  }
  \label{fig:feature_fusion}
\end{figure}

\noindent{\textbf{Visualization of the original document data features sampled by different methods.} }
In addition, to evaluate the feasibility of direct feature extraction from raw instructional document data to sample effective instances, the widely acknowledged feature extractor CLIP \cite{radford2021learning} is utilized to extract features from images and corresponding instructional texts within document datasets gathered by various sampling methods. Subsequently, these features are visualized using t-SNE \cite{arora2018analysis}.
As depicted in Figure~\ref{fig:feature_fusion}, the raw document instruction features originating from diverse sample sources could not be effectively differentiated, despite observable performance variances amongst them. This indicates that for document instruction data, utilizing raw data features does not afford effective discrimination. Hence, it is imperative to employ specialized document representations like modeling the process of document instruction execution for effective modeling rather than relying on raw data.

\subsection{ProcTag for Data Efficacy}
Existing instruction datasets employed for LLMs/MLLMs are often sourced from generated content, characterized by significant size and redundancy, which can lead to extended training durations and resource inefficiency.
To address this challenge, data efficacy is introduced, denoted as $E$, to serve as a criterion for selecting high-quality data. 
This efficacy is defined by the ratio of current performance $P_{cur}$ to the best performance $P_{best}$, given by $E = P_{cur} / P_{best}$.
To assess data efficacy, we conduct experiments based on data ratio and tag coverage rate. Within these two settings, $P_{\text{best}}$ refers to the performance achieved at 100\% data ratio and tag coverage rate, respectively.

As shown in Figure~\ref{fig:efficiency_200k}, when the data ratio is varied, the peak performance is observed at 30\%, indicating that our ProcTag can maintain high effectiveness with only a small subset of data. 
Furthermore, in terms of coverage rate, which represents the percentage of all tags covered in the dataset.
Experimental results show that as coverage increases, performance exhibits a positive correlation with improvement. The results confirm the effectiveness of all tags generated by ProcTag, indicating that each tag contributes to the enhancement of performance. These results validate that our ProcTag for tagging is rational and effective, while also ensuring the diversity and complexity of these tags.

\begin{table}
  \centering
  \begin{tabular}{ccc}
    \hline
    Verbalizer & \begin{tabular}[c]{@{}c@{}}GPT-3.5\\ (direct prompting)\end{tabular} & \begin{tabular}[c]{@{}c@{}}Qwen\\ (fine-tuning)\end{tabular} \\ \hline
    PlainText & 63.66 & 77.14 \\
    SpatialFormat & 71.03 & 80.93 \\
    DocLayPrompt & \textbf{73.84} & \textbf{81.80} \\ \hline
    \end{tabular}
    \label{tab:ablation_doclayprompt}
    \caption{
    Ablation study on the impact of different prompts for representing documents in the DocVQA task.
      }
  \end{table}

\subsection{Ablation study}
\subsubsection{Impact of DocLayPrompt}

  \begin{figure}
    \centering
    \includegraphics[width=0.9\linewidth]{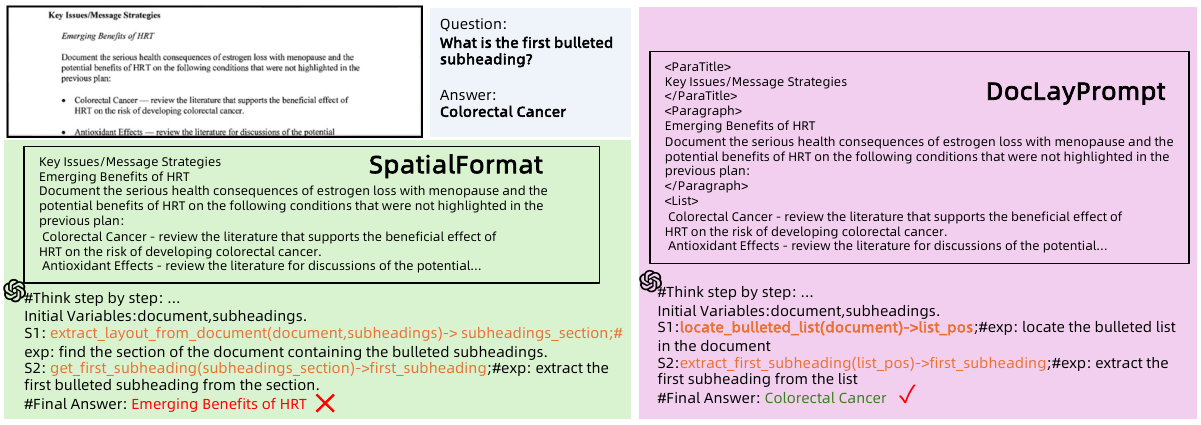}
    \caption{
    Case study on document instruction execution process generation using different document representations.
    }
    \label{fig:aba_doclayprompt_casestudy}
  \end{figure}


To evaluate the effectiveness of DocLayPrompt in document text representation, we conducted comparative experiments with two commonly used prompts: PlainText and SpecialFormat~\cite{lamott2024lapdoc}. In direct prompting tests on the DocVQA test set using GPT-3.5, as shown in Table~\ref{tab:ablation_doclayprompt}, DocLayPrompt improved ANLS by 2.81 compared to SpecialFormat~\cite{lamott2024lapdoc}. Additionally, fine-tuning experiments with Qwen using various document prompts showed that DocLayPrompt outperformed both PlainText and SpecialFormat, demonstrating its superiority in document representation, whether for direct prompting or LLM training.


Figure~\ref{fig:aba_doclayprompt_casestudy} shows the differences in the document instruction process generation using SpatialFormat and DocLayPrompt.
Employing SpatialFormat prompts in the instruction execution generation failed to guide GPT in generating answers consistent with the accurate annotations of the original instruction dataset.
It is observed that GPT fails to recognize key layout information from such textual representation, thus not generating tags related to the list-region.
By integrating layout information via DocLayPrompt for document representation, the instruction execution generation process produced final answers that aligned with the annotations in the original instruction dataset. 
And it can make GPT understand the document layout, like \textit{locate\_bulleted\_list}.
This underscores the efficacy of the generated instruction execution process.

\subsubsection{Impact of different representations of the process.}

  \begin{figure}[tb]
  \centering
  \includegraphics[width=0.75\linewidth]{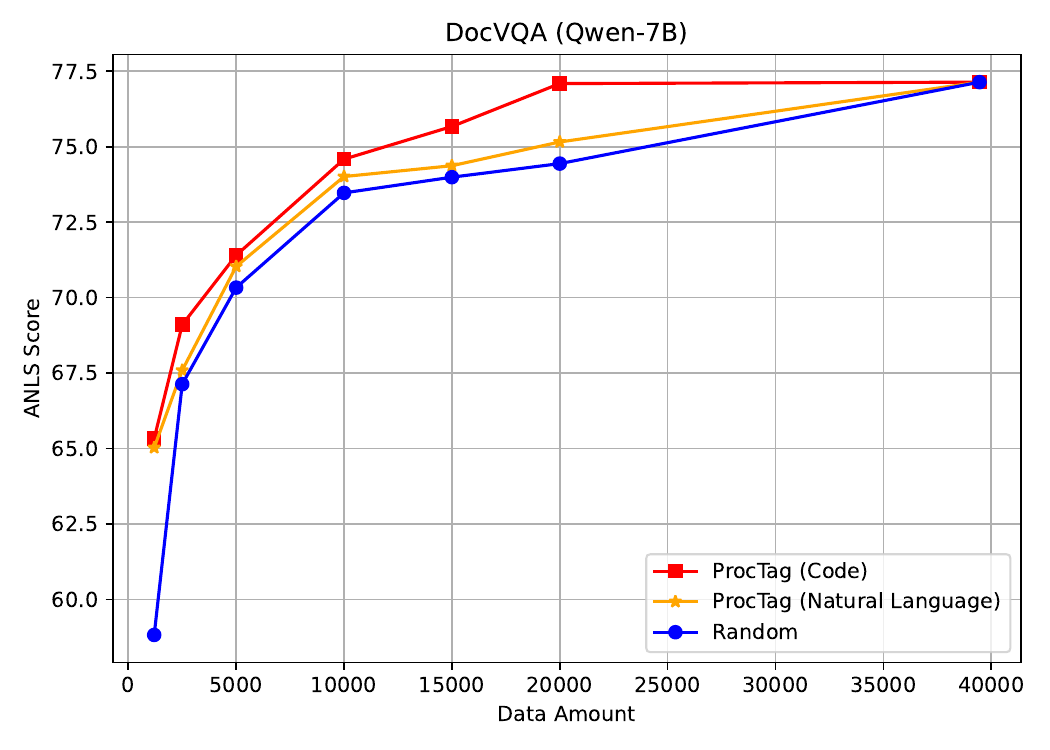}
  \caption{The impact of representing instruction execution processes with pseudo-code and natural language.
  }
  \label{fig:ablation_code_text}
\end{figure}
Though a straightforward process representation using natural language is recognized as a practical alternative, pseudo-code is utilized to represent the instruction execution process in ProcTag.
The preference for pseudo-code stems from its ability to provide a more granular and structured representation, which is anticipated to yield a more detailed and succinct explanation of the instruction execution process. Thus, it is likely to enhance the effectiveness of the data sampling process.
To evaluate the effectiveness of pseudo-code representation, an analysis is conducted with process tags generated in natural language and assessed using the DocVQA dataset. 
As shown in Figure~\ref{fig:ablation_code_text}, the experimental results show that while natural language tags offer some advantages over random sampling, they do not provide the significant improvements observed with pseudo-code tags, confirming the superior efficacy of the pseudo-code format.

\section{Limitation}

Due to cost constraints, the current approach employs GPT-3.5 and a text-based document prompt,  rather than incorporating multimodal large language models like GPT-4V.
Clearly, relying on textual document representation results in the loss of certain visual information, thereby hindering the applicability of our approach on visually-rich datasets such as InfographicVQA~\cite{mathew2022infographicvqa}.

\section{Conclusion}
In this paper, we propose ProcTag, a data method that assesses the efficacy of document instruction data.
ProcTag performs tagging on the execution process of document instructions and utilizes the diversity and complexity of these tags to assess the efficacy of the dataset.
Additionally, DocLayPrompt, a novel semi-structured layout-aware document prompt, is proposed for effectively representing documents.
Experimental results demonstrate the effectiveness of the ProcTag method in assessing document instruction data with efficacy when compared to existing data methods and random sampling.
As modeling the process of instruction execution is a generic approach, in the future, we will extend this approach to the general artificial intelligence domain, exploring more effective data evaluation strategies.

\section{Acknowledgments}
This work is supported by the National Natural Science Foundation of China (Grant No. 62372408)


\bibliography{aaai25}

\end{document}